\documentclass[runningheads]{llncs}
\usepackage{graphicx}
\usepackage{tikz}
\usepackage{comment}
\usepackage{amsmath,amssymb} % define this before the line numbering.
\usepackage{color}

\usepackage{enumitem}
\usepackage{algorithm}
\usepackage{algpseudocode}
\usepackage{multirow}
\usepackage{graphicx}
\usepackage{svg}
\usepackage{float}
\usepackage[accsupp]{axessibility}  % Improves PDF readability for those with disabilities.

\begin{document}
% \linenumbers
\pagestyle{headings}
\mainmatter
\def\ECCVSubNumber{177}  % Insert your submission number here

\title{Spacecraft Pose Estimation Based on Unsupervised Domain Adaptation and on a 3D-Guided Loss Combination} % Replace with your title

\begin{comment}
\titlerunning{ECCV-22 submission ID \ECCVSubNumber} 
\authorrunning{ECCV-22 submission ID \ECCVSubNumber} 
\author{Anonymous ECCV submission}
\institute{Paper ID \ECCVSubNumber}
\end{comment}
%******************

%\begin{comment}
\titlerunning{Spacecraft Pose Estimation Based on UDA and 3D-Guided Loss}
\author{Juan Ignacio Bravo Pérez-Villar\inst{1,2}\index{Bravo Pérez-Villar, Juan Ignacio} \and
Álvaro García-Martín\inst{2} \and
Jesús Bescós\inst{2}}
\authorrunning{J. I. Bravo Pérez-Villar et al.}
\institute{Deimos Space, 28760, Madrid, Spain\\ \email{juan-ignacio.bravo@deimos-space.com} \and
Video Processing and Understanding Lab, Univ. Autónoma de Madrid, 28049, Madrid, Spain\\ \email{juanignacio.bravo@estudiante.uam.es} \\\email{\{alvaro.garcia, j.bescos\}@uam.es}
}
%\end{comment}
%******************

\maketitle

%-------------------------------------------------------------------------
%  Abstract           
%-------------------------------------------------------------------------

\begin{abstract}
Spacecraft pose estimation is a key task to enable space missions in which two spacecrafts must navigate around each other. Current state-of-the-art algorithms for pose estimation employ data-driven techniques. However, there is an absence of real training data for spacecraft imaged in space conditions due to the costs and difficulties associated with the space environment. This has motivated the introduction of 3D data simulators, solving the issue of data availability but introducing a large gap between the training (source) and test (target) domains. We explore a method that incorporates 3D structure into the spacecraft pose estimation pipeline to provide robustness to intensity domain shift and we present an algorithm for unsupervised domain adaptation with robust pseudo-labelling. Our solution has ranked second in the two categories of the 2021 Pose Estimation Challenge organised by the European Space Agency and the Stanford University, achieving the lowest average error over the two categories\footnote{Code is available at: \url{https://github.com/JotaBravo/spacecraft-uda}}.

\keywords{Spacecraft, Pose Estimation, Unsupervised Domain Adaptation, 3D Loss.}
\end{abstract}

%-------------------------------------------------------------------------
%  Introduction           
%-------------------------------------------------------------------------

\section{Introduction}

Relative navigation between a chaser and a target spacecraft is a key capability for future space missions, due to the relevance of close-proximity operations within the realm of active debris removal or in-orbit servicing. Camera sensors are a growing suitable choice to aid in the relative navigation around non-cooperative targets due to their reduced cost, low-mass and low-power requirements compared to active sensors. However, data acquired in real operational scenarios is often scarce or not available, limiting the deployment of data-driven algorithms. This limitation is nowadays overcome by using data simulators, which solve the issue of data scarcity but introduce the problem of domain shift between the training and test datasets.

The 2021 Spacecraft Pose Estimation Challenge (SPEC2021) \cite{kelvins}, organised by the European Space Agency and the Stanford University, was designed to bridge the performance gap between computer-simulated and operational data for the task of non-cooperative spacecraft pose estimation. SPEC2021 builds around the SPEED+ dataset \cite{park2021speedplus} containing 60,000 computer-generated images of the Tango spacecraft with associated pose labels and 9531 unlabelled  hardware-in-the-loop test images of a half-scale mock-up model. Test images are grouped into two subsets: Sunlamp, consisting of 2791 images featuring strong illumination and reflections over a black background; and Lightbox, consisting of 6740 images with softer illumination but increased noise levels and the presence of the Earth in the background (see sample images in Fig. \ref{fig:speed_plus_dataset}). SPEC2021 evaluates the accuracy in pose estimation for each test subset. 

Under this Challenge, we have developed a single algorithm that has ranked second in both Sunlamp and Lightbox categories, with the best total average error over the two datasets. Our main contributions are:
\begin{itemize}
    \item A spacecraft pose estimation algorithm that incorporates 3D structure information during training, providing robustness to intensity based domain-shift. 
    \item An unsupervised domain adaptation scheme based on robust pseudo-label generation and self-training.
\end{itemize}

\begin{figure}[t]
\begin{center}
   \includegraphics[width=1\linewidth]{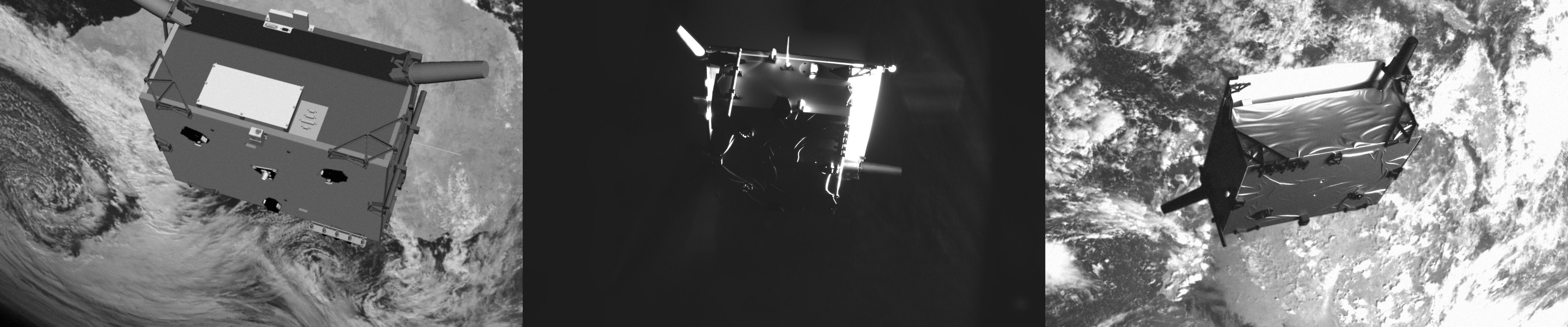}
\end{center}
   \caption{Sample images of the SPEED+ Dataset. From left to right: Synthetic (training \& validation), Sunlamp (test), Lightbox (test).}
\label{fig:long}
\label{fig:onecol}
\end{figure} \label{fig:speed_plus_dataset}

%-------------------------------------------------------------------------
%  Related Work           
%-------------------------------------------------------------------------

\section{Related Work}
 This section briefly presents the related  work on pose estimation and unsupervised domain adaptation.
\subsection{Pose Estimation}

Pose estimation is the process of estimating the rigid transform that aligns a model frame with a sensor, body, or world reference frame \cite{kelsey2006vision}. Similarly to \cite{peng2019pvnet} we categorise these methods into: end-to-end, two-stage, and dense. 

\textbf{End-to-end} methods map the two-dimensional images or their feature representations into the six-dimensional output space without employing geometrical solvers. Handcrafted approaches compare image regions with a database of templates to return the 6 Degree-of-freedom (DoF) pose associated with the template \cite{hinterstoisser2010dominant,shi2017spacecraft}. Learning-based approaches employ CNNs to either learn the direct regression of the 6DoF output space \cite{kendall2015posenet,mahendran20173d,kendall2017geometric,proencca2020deep} or classify the input into a discrete pose \cite{su2015render}. End-to-end methods have been explored in \cite{sharma2018pose}, where an AlexNet was used to classify input images with labels associated to a discrete pose. The authors in \cite{proencca2020deep} employed a network to regress the position and do continuous orientation estimation of the spacecraft with soft assignment coding. 

\textbf{Two-stage} pose estimation methods first estimate the 2D image projection of the 3D object key-points and then apply a Perspective-n-Point (PnP) algorithm over the set of 2D-3D correspondences to retrieve the pose. Traditionally, this has been solved via key-point detection, description and matching between the images and a 3D model representation based on handcrafted \cite{lowe2004distinctive,bay2008speeded,alcantarilla2012kaze} or learnt \cite{sun2021loftr,mishchuk2017working} key-point detectors, descriptors and matchers. Other approaches directly employ CNNs designed for 2D key-point localisation that detect the vertices of the 3D object bounding-box or regress the key-point position by employing heatmaps, hence unifying the detection, description and matching into a single network \cite{newell2016stacked,wang2020deep}. These approaches are the current state-of-the-art in satellite pose estimation. The authors from \cite{chen2019satellite} employed an HRNet and non-linear pose refinement which obtained the first position in the Spacecraft Pose Estimation Challenge of 2019. In  \cite{hu2021wide} a multi-scale fusion scheme with a 3D loss independent of the depth was employed to improve the pose estimation accuracy under different scale distributions of the spacecraft.

\textbf{Dense} pose estimation methods establish dense correspondences, at pixel or patch level, between a 2D image and a 3D surface object representation. These approaches can be based on depth data, RGB images or image pairs \cite{peng2019pvnet,kehl2016deep,doumanoglou2016recovering}.

\subsubsection{Discussion:}
Non-cooperative spacecraft pose estimation presents specific conditions that can be incorporated as prior knowledge into the algorithms: 1) the target will not appear occluded in the image; 2) and the targets have highly reflective materials that coupled with small orbital periods frequently lead to heterogeneous illumination conditions during rendezvous; and 3) the estimation should provide some degree of interpretability so the navigation filter can handle failure cases. End-to-end pose estimation methods, that confront the problem as a classification task, are limited by the error associated with the discretisation of the pose-space and require a correct sampling of the attitude space to obtain uniform representations \cite{kuffner2004effective}. In addition, the methods that directly regress the pose fail to generalise, as the formulation might be closer to the image retrieval problem rather than to pose estimation one \cite{sattler2019understanding}. Two-stage pose estimation methods do not provide robustness to occlusions and can be affected by key-points falling outside the image space. Finally, dense methods provide robustness to occlusions by producing a prediction for every pixel or patch in the image, but increase the difficulty of the learning task due to the larger output space \cite{peng2019pvnet}. These also impose additional constraints on the input data as methods may need depth or surface normals, which were not available for the Challenge.

These considerations motivated our choice of a two-stage approach. Key-points provide a certain degree of interpretability, as the output is easily understandable and can be directly related with a pose. These methods also allow for continuous output estimation without compromising generalisation and do not require the additional ground-truth data, usually depth, that dense methods do. In addition, two-stage methods have shown better performance compared to end-to-end methods in 6DoF pose estimation \cite{shavit2019introduction} and are commonly used into the state-of-the-art methods for the spacecraft domain \cite{hu2021wide,chen2019satellite}.

\subsection{Domain Adaptation}
In visual-based navigation for space applications the distribution of the data to be acquired during the mission is often unknown, as no or little data is available prior the mission. This entails the introduction of computer graphic simulators \cite{parkes2004planet} to generate the data required to design and test the algorithms. Although computer-graphic simulated images have greatly improved their quality and fidelity, there still exists a noticeable domain gap between the simulated (source) data and the real (target) data. Methods to overcome the reduction in algorithms performance for the same task under data with different distributions fall under the category of Domain Adaptation \cite{pan2009survey}. More specifically, when labels are only provided in the source domain is referred as Unsupervised Domain Adaptation (UDA). UDA methods can be categorised into: 1) domain-invariant feature learning; 2) aligning input data; and 3) self-training via pseudo-labelling. 

\textbf{Domain-invariant feature learning} aims to align the source and target domain at feature level. It is based on the assumption that exists a feature representation that does not vary across domains, i.e. the information captured by the feature extractor is the one that does not depend on the domain. This family of methods try to learn the representation either by minimising the distribution shift under some divergence metric \cite{sun2016deep}, enforcing that both feature representations can be used to reconstruct the target or source domain data \cite{ghifary2016deep}, or employing adversarial training \cite{ganin2015unsupervised}. 

\textbf{Input data alignment}. Instead of aligning the domains at the feature level, these approaches aim to align the source domain to the target domain at input level, a mapping usually confronted via generative adversarial networks \cite{zhu2017unpaired,shrivastava2017learning}. 

\textbf{Self-training} approaches employ generated pseudo-labels to iteratively improve the model. However, these labels generation process is inevitably noisy. Some works have tried to mitigate the effect of the label noise by designing robust losses \cite{zhang2018generalized} but they often fail to handle real-world noisy data \cite{zhang2021prototypical}. Other approaches have used self-label correction that relies on the agreement between different learners \cite{lee2018cleannet} or stages of the pipeline \cite{zhang2021prototypical} to correct the noisy labels. Online domain adaptation is employed in \cite{park2022robust} for satellite pose estimation across different domains. A segmentation-based loss is employed to update the feature extractor of a network, adapting the features to the new input-domain.

\subsubsection{Discussion:}
Feature learning methods rely on the existence of feature representations that do not vary across domains, and assume that similar semantic objects are shared between domains during training. However, in pose estimation the semantic objects (key-points) are pose dependent and thus, the same semantic objects at training might not appear in both domains, hence affecting the alignment performance. Input alignment methods are able to transfer general illumination and contrast properties. However, they might introduce artefacts, and reflectances due to the object materials and sensor specific noise between train and test are often not captured. Our proposal is based on pseudo-labelling, which allows to learn from the target domain providing directly interpretable results. In addition, pseudo-labelling provides state-of-the-art results in UDA for the semantic segmentation task \cite{zhang2021prototypical}.

%-------------------------------------------------------------------------
%  Pose Estimation Method    
%-------------------------------------------------------------------------
\section{Pose Estimation} \label{section:PoseEstimation}

\begin{figure}[t]
\begin{center}
   \includegraphics[width=1\linewidth]{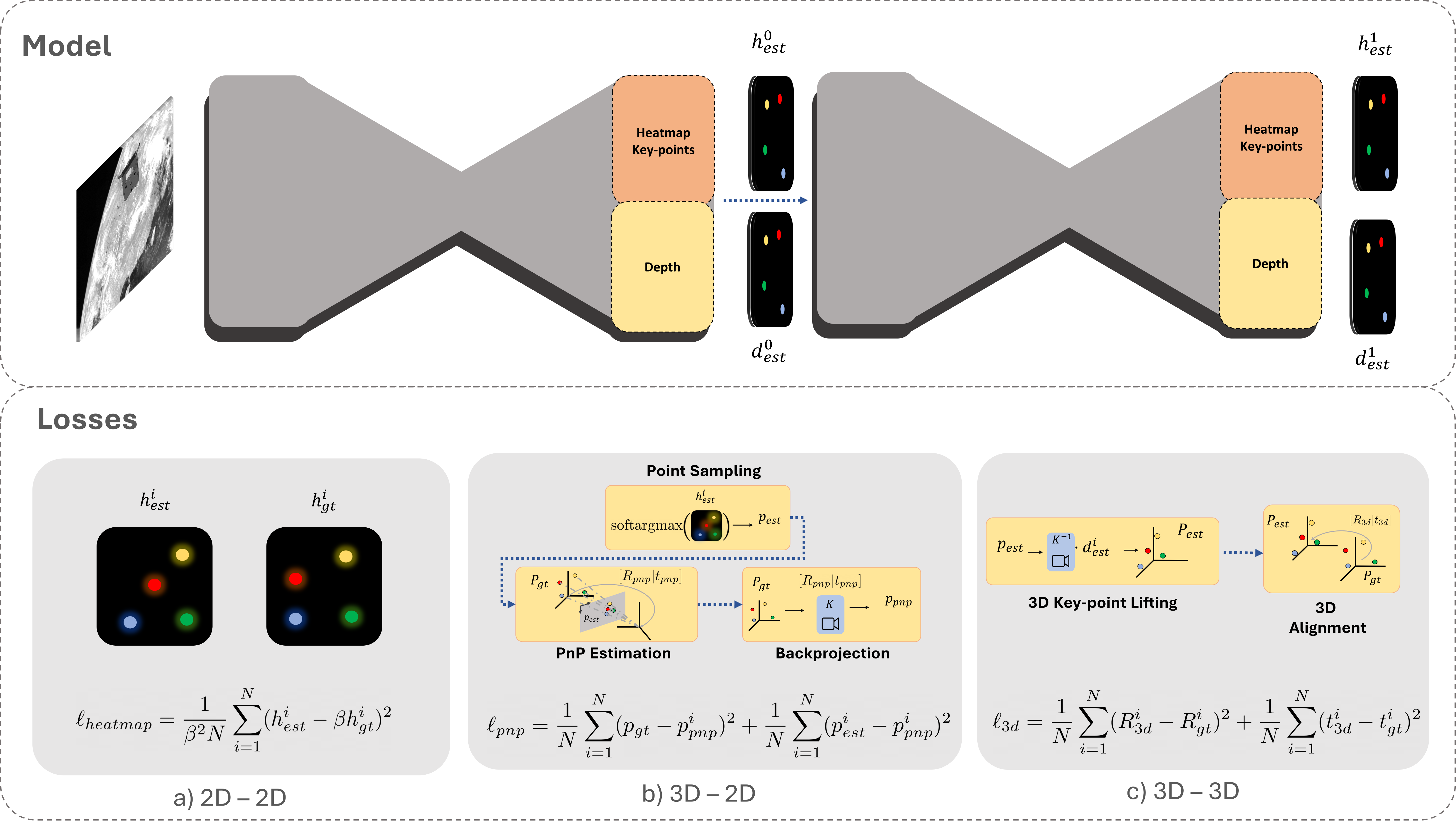}
\end{center}
   \caption{Proposed pipeline. Model (top) - A two-stack hourglass network with two heads for key-point extraction and depth prediction. Losses (bottoom) -  A combination of: a) 2D-2D comparison between the predicted key-point heatmaps and the ground-truth heatmaps; b) 2D-3D comparison between the 3D key-point positions estimated via PnP and the ground-truth positions; c) 3D-3D comparison between the pose that aligns the lifted 2D estimated key-points with the 3D model and the ground-truth pose.}
\label{fig:architecture}
\end{figure}

This section introduces the proposed method for pose estimation. In our setting, the target frame is the geometrical centre of the target spacecraft, represented by a set of 11 3D key-points. The sensor frame is expressed at the centre of the image. At test time, we use a Perspective-n-Point solver over the 3D spacecraft key-points and their estimated 2D image projections to retrieve the pose.

In this challenging scenario where, apart from a noticeable domain gap, surfaces reflectivity and illumination conditions further generate great change, objects in the source and target domains keep the same 3D structure. This motivates the use of an architecture and learning losses that not only rely on image textures, but make explicit use of the 3D key-point structure. 

At architecture level (see Fig. \ref{fig:architecture} - Model), we propose the use of a stacked hourglass network with two heads to jointly regress the key-point position and their associated depth, enabling 3D-aware feature learning. At loss level (see Fig. \ref{fig:architecture} - Losses), we formulate a triple learning objective: a) in the 2D-2D loss we compare the mean squared error between the key-point head output with a 2D Gaussian blob centred on the ground-truth key-point location; b) in the 3D-2D loss we incorporate pose information by comparing the coordinates of the 3D key-points projected onto the image using the estimated pose via PnP with the corresponding ground-truth coordinates in the image; c) in the 3D-3D loss we incorporate 3D key-point structure: the estimated depth along the camera intriniscs is used to lift the estimated key-point positions, and these lifted key-points are aligned in 3D with the ground-truth key-points using a least square estimator. Then, the retrieved pose from the alignment is compared to the ground-truth pose.

\subsection{2D-2D: Key-point Heatmap Loss}\label{seq:pnp_loss}

The ground-truth heatmaps $h_{gt}$ are obtained by representing a circular Gaussian centred at the ground-truth image key-point coordinates $p_{gt} = [u_{gt}, v_{gt}]^T$. Given an image of the target spacecraft, represented by the 3D key-points $P_{gt} = [x,y,z]^T$ and acquired with camera intrinsics $K$ in a ground-truth pose with rotation matrix $R_{gt}$ and translation vector $t_{gt}$, the ground-truth key-point coordinates can obtained via the projection expression:

\begin{equation}
        p_{gt} = KR_{gt}^T(P_{gt} + R_{gt}t_{gt}).\label{eq:projection_2d}
\end{equation}

The heatmaps $h^i_{est}$ predicted by each head $i$ are learnt by comparing the output of the key-point prediction head vs the ground-truth heat-map $h_{gt}$ via mean squared error. To keep a fixed range for the key-point heatmap loss, we normalise each heatmap by its maximum value and weight them by a parameter $\beta=1e3$, ensuring a large difference between the minimum and maximum value of the heatmap. The key-point heatmap loss is then computed via:

\begin{equation}
    \ell_{heatmap} = \frac{1}{\beta^2N}\sum_{i=1}^N(h^i_{est} - \beta h^i_{gt})^2.
\end{equation}\label{eq:loss_heatmap}

\subsection{3D-2D: PnP Loss}

Pose information is here considered by comparing the ground-truth 2D key-point image coordinates $p_{gt}$ with the projection of the 3D key-points $P_{gt}$ onto the image, obtained using the rigid transform estimated via PnP (see Fig. \ref{fig:architecture} - Model b)). The predicted key-point locations $p^i_{est}$ used to estimate such rigid transform are originally expressed as heatmaps. The application of the arg-max function over $h^i_{est}$ to retrieve the predicted key-point pixel positions $p^i_{est}$ is a non-differentiable operation which cannot be used for training. We instead use the integral pose regression method \cite{sun2018integral} and apply an expectation operation over the soft-max normalized heatmaps $h^i_{est}$ to retrieve the pixel coordinates $p^i_{est}$. Then, we employ the backpropagatable PnP algorithm from \cite{chen2020end} to retrieve the estimated rotation matrix $R^i_{pnp}$ and translation vector $t^i_{pnp}$. 

The PnP loss evaluates the mean squared error between the ground-truth key-point pixel locations $p_{gt}$ and the key-point pixel locations $p^i_{pnp}$ obtained by projecting $P_{gt}$ onto the image with $R^i_{pnp}$ and $t^i_{pnp}$ as in Eq. \ref{eq:projection_2d}. The loss has an additional regularisation term, comparing $p_{est}$ with $p_{pnp}$, to ensure convergence of the estimated key-point coordinates the desired positions \cite{chen2020end}. The final PnP loss term $\ell_{pnp}$ is computed via:

\begin{equation}
    \ell_{pnp} = \frac{1}{N}\sum_{i=1}^N(p_{gt}-p^i_{pnp})^2 + \frac{1}{N}\sum_{i=1}^N(p^i_{est}-p^i_{pnp})^2.
\end{equation}\label{eq:loss_pnp}

\subsection{3D-3D: Structure Loss} 
We incorporate 3D structure by comparing the rotation $R^i_{3d}$ and translation $t^i_{3d}$ that align the predicted key-points lifted to the 3D space $P^i_{3d}$ with the key-points representing the spacecraft $P_{gt}$ with the ground-truth pose (see Fig. \ref{fig:architecture} - Model c)). Following the approach described in Section \ref{seq:pnp_loss}, we retrieve the estimated key-point coordinates $p^i_{est}$. Then we lift them to the 3D space by applying the camera intrinsics $K$ and the estimated depth $d^i_{est}$. Instead of providing direct supervision to the depth, we choose to sample the depth prediction head at the position of the ground-truth key-point positions $p^i_{gt}$, and supervise the output with the 3D alignment loss. Sampling at the ground-truth key-point positions instead of the estimated positions helps convergence, as the number of combinations of key-point locations and depth values that satisfy a given rotation and translation is not unique. $R^i_{3d}$ and $t^i_{3d}$ are found by minimising the sum of square distances between the target points $P_{gt}$ and the lifted 3d points $P^i_{3d}$ :

\begin{equation}
    R^i_{3d}, t^i_{3d} = \mathtt{argmin}_{R,t} \sum \|R P^i_{3d} + t - P_{gt}\|^2
\end{equation} \label{eq:3d}

The algorithm that solves this expression is referred as the Umeyama \cite{umeyama1991least} (or  Kabsch) algorithm and has a closed form solution. We employ the Procrustes solver from \cite{bregier2021deepregression} to obtain the arguments that minimise the above expression. The loss is then computed by:

\begin{equation}
    \ell_{3d} = \frac{1}{N}\sum_{i=1}^N(R^i_{3d}-R^i_{gt})^2 + \frac{1}{N}\sum_{i=1}^N(t^i_{3d}-t^i_{gt})^2
\end{equation}

The final learning objective is then defined by a weighted sum of the three losses. The values of $\gamma_{1} = \gamma_{2} = 0.1$ are selected so the contributions of the 2D-3D and 3D-3D losses represent a 10\% of the total learning objective.

\begin{equation}
    \ell_{final} = \ell_{heatmap} + \gamma_{1}\ell_{pnp} + \gamma_{2}\ell_{3d}.
\end{equation}

\section{Domain Adaptation}\label{sec:domain_adaptation} 

The losses defined in Section \ref{section:PoseEstimation} to consider texture,  pose and structure improve the performance in the test set for the task of spacecraft pose estimation (see Section \ref{seq:additional_experiments}). However,  we still observe a large performance gap between the source domain (validation) and the two target domains (test). This gap motivates the introduction of unsupervised domain adaptation techniques. Given a dataset of images in the source domain $I_s$ and the associated ground-truth rotation matrix $R_{gt}$ and translation vector $t_{gt}$ that define the pose, we aim to train a model $M$ such that when evaluated on a target dataset $I_t$ produces correct ground-truth key-point positions, without using the ground-truth pose. Our proposal relies on self-training with robust pseudo-labelling generation. As discussed before, pseudo-labels provide a strong supervision signal on the target domain but are intrinsically noisy. If not handled properly, this noise can degrade the overall performance. In our setting, we reduce the noise during label generation by enforcing consistent pose estimates across the network head.

The proposed approach is summarised in Algorithm \ref{alg:algorithm1}. We first pre-train a model $M$ (Fig. \ref{fig:architecture} - Model) on the source dataset. Then, we evaluate the images from a given target dataset $I_t$ on the model $M$. The model $M$ will generate a prediction on the key-point locations, expressed as heatmaps $h^i_{est}$ at each head $i = 1,2$ of the network. We consider the pixel coordinates of the maximum value of the heatmap to be the predicted key-point pixel coordinates $p^i_{est}$. Differently to the optimisation of the losses $\ell_{pnp}$ and $\ell_{3d}$, here we do not require this step to be differentiable, and thus a standard argmax operation suffices: $p^i_{est} = argmax(h^i_{est})$. Next, we perform key-point filtering based on the heatmap response. First, we remove the key-points falling close to the edges of the image. The remaining key-points $p^i_{est}$ are sorted based on the total sum of the heatmap $h^i_{est}$, and the seven key-points with strongest response are kept. The remaining key-points are noted as $p^i_{filt}$. We employ the the EPnP (Efficient Perspective-n-Point) algorithm \cite{lepetit2009epnp} under a RANSAC loop to estimate the pose with the filtered key-points $p^i_{filt}$ at each head $i$.

The pose at each level is estimated with the prediction of each head  $p^i_{filt}$ by applying the EPnP (Efficient Perspective-n-Point) algorithm \cite{lepetit2009epnp} under a RANSAC loop. A new pose pseudo-label is introduced into the test set if the RANSAC-EPnP routine converges for all the heads $i$. The solution corresponding to the largest total key-point response is chosen as the resulting pose. The convergence of the RANSAC routine is controlled by two parameters: the confidence parameter $c$ that controls the probability of the method to provide a useful result, and the reprojection error $r$ that defines the maximum allowed distance between a point and its reprojection with the estimated pose to be considered an inlier. Modifying the convergence criteria (i.e. $c$ and $r$) allows to control the amount of label noise. In our case, we choose $c=0.999$ and $r=2.0$. 
\begin{algorithm}[h]
\caption{Pseudo-Labelling for Spacecraft Pose Estimation.}
\begin{algorithmic}[1]
\While{convergence criteria}
    \If{first iteration}
    \State Initialise $M$ with weights trained on source domain 
    \Else
    \State Initialise $M$ with weights from previous iteration
    \EndIf
    \For{$j = 1$ to $J$}
        \State $h^i_{est} = M(I^j_t)$ 
        \For{$i = 1:2 $} 
            \State $p^i_{est}  = \textrm{argmax}(h^i_{est})$
            \State $p^i_{filt} = \textrm{FilterKeypoints} (p^i_{est})$ 
            \State $R^i_{est}, t^i_{est}, \textrm{flag}^i = \textrm{PoseEstimation} (p^i_{filt}, P_{gt},c=0.999,r=2.0)$ 
        \EndFor
        \If{$\textrm{flag}^i == \textrm{True}$ for all i}
            \State Choose $R_{est}, t_{est}$ from key-point response
            \State Add $R_{est}, t_{est}$ pseudo-label to $I^j_t$
        \EndIf
    \EndFor
    \State Retrain $M$ for a number of epochs 
\EndWhile
\end{algorithmic} \label{alg:algorithm1}
\end{algorithm}

%-------------------------------------------------------------------------
%  Pose Estimation Method    
%-------------------------------------------------------------------------
\section{Evaluation}
%In this section we present the evaluation metrics and challenge results, including image representation of our results. Additionally we show some experiments.

\subsection{Metrics}
The evaluation metric used in the Challenge was the pose score $S_{total}$. The pose score is based on the combination of the position score $S_{pos}$ and the orientation score $S_{ori}$. The position score for an image $j$ is defined as the absolute difference between the estimated and ground-truth translation vectors, normalised by the module of the ground-truth position vector. The position score accounts for the precision in position of the robotic platform, which is of $2.173$ millimetre per metre of ground truth distance. Values of $S_{pos}$ lower than $0.002173$ are zeroed. 

\begin{equation}
    S^j_{pos} = \frac{\left|t^j_{est} -  t^j_{gt} \right|_2}{\left|t^j_{gt}\right|_2}
\end{equation}

%\begin{equation}
%    S^j_{pos} = 
%   \begin{cases}
%         E^j_{pos} = \frac{\left|t^j_{est} -  t^j_{gt} \right|_2}{\left|t^j_{gt}\right|_2} & \textrm{if} E^j_{pos} < 0.002173\\
%         E^j_{pos} & \textrm{otherwise}
%    \end{cases}
%\end{equation}

The orientation score $S_{ori}$ is defined as the angle that aligns the estimated and ground-truth quaternion orientations. This metric also accounts for the precision of the robotic arm which is 0.169º, zeroing $S_{ori}$ than the precision.

\begin{equation}
  S^j_{ori} = 2\cdot arccos(\left| \langle q^j_{est},q^j_{gt}\rangle\right|)
\end{equation}

%\begin{equation}
%    S^j_{ori} =
%    \begin{cases}
%        0 & \textrm{if}E^j_{ori} < 0.169^{\circ}\\
%       E^j_{ori} & \textrm{otherwise}
%   \end{cases}
%\end{equation}

The total score is the average of the sum of the orientation and position scores corresponding the $N$ images of the test set:

\begin{equation}
    S_{total}^j = \frac{1}{J} \sum^J_{j=1} ( S^j_{pos} + S^j_{ori} )
\end{equation}

%\begin{equation}
%    score^{i}_{\textrm{position}} = 
%    \begin{cases}
%%%        0 & \textrm{if err^{i}_{\textrm{position}}} < $0.002173 mm/m$ \left| r^i_{gt}\right|_2 \\
%        err^i__{\textrm{position}} & \textrm{otherwise}
%    \end{cases}
%\end{equation}

\subsection{Challenge results}

The ground-truth labels of the two test sets (Lightbox and Sunlamp) from the SPEED+ dataset have not been made publicly available. Instead, the Challenge evaluation was performed by submitting the estimated poses to a evaluation server; the results were then published on public leaderboards only if the obtained score improved a previous result from the same participant. We provide a summary of the results by showing the error achieved by the 10 first teams on the Lightbox and Sunlamp datasets on Table \ref{tab:leaderboard_lightbox} and Table \ref{tab:leaderboard_sunlamp} respectively. Teams are represented in decreasing order based on the achieved rank in the leaderbord, corresponding the first row to the winner of the category. The winning teams were \textit{TangoUnchained} for the Lightbox dataset and \textit{lava1302} for the Sunlamp dataset. Our solution \textit{VPU} ranked 2nd on both datasets, and achieved the best total average error as shown in Table \ref{tab:leaderboard_average}, where we represent the averaged scores of the two datasets. This shows that our solution is robust to different domain changes and can be applied without fine-tuning on different settings with different illumination and noise conditions. 

\begin{table}[H]
\centering
\resizebox{0.65\textwidth}{!}{%
\begin{tabular}{l|cc|c}
\multicolumn{1}{c|}{\textbf{Team Name}} & \textbf{$S^j_{pos}$} & \textbf{$S^j_{ori}$} & \textbf{ $S^j_{total}$} \\ \hline
TangoUnchained                          & \textbf{0.0179 }     & \textbf{0.0556}    & \textbf{0.0734}     \\
VPU  (Ours)         & 0.0215 & 0.0799 & 0.1014 \\
lava1302            & 0.0483 & 0.1163 & 0.1646 \\
haoranhuang\_njust  & 0.0315 & 0.1419 & 0.1734 \\
u3s\_lab            & 0.0548 & 0.1692 & 0.2240 \\
chusunhao           & 0.0328 & 0.2859 & 0.3186 \\
for graduate        & 0.0753 & 0.4130 & 0.4883 \\
Pivot SDA AI \& Autonomy Sandbox        & 0.0721              & 0.4175             & 0.4895              \\
bbnc                & 0.0940 & 0.4344 & 0.5312 \\
ItTakesTwoToTango   & 0.0822 & 0.5427 & 0.6248 
\end{tabular}%
}
\caption{Summary of the SPEC2021 results for the Lightbox Dataset.}
\label{tab:leaderboard_lightbox}
\end{table}

% \usepackage{graphicx}
\begin{table}[htbp]
\centering
\resizebox{0.65\textwidth}{!}{%
\begin{tabular}{l|cc|c}
\multicolumn{1}{c|}{\textbf{Team Name}} & \textbf{$S^j_{pos}$} & \textbf{$S^j_{ori}$} & \textbf{ $S^j_{total}$} \\ \hline
lava1302                           & \textbf{0.0113} & \textbf{0.0476} & \textbf{0.0588} \\
VPU (Ours)                         & 0.0118          & 0.0493          & 0.0611          \\
TangoUnchained                     & 0.0150          & 0.0750          & 0.0899          \\
u3s\_lab                           & 0.0320          & 0.1089          & 0.1409          \\
haoranhuang\_njust                 & 0.0284          & 0.1467          & 0.1751          \\
bbnc                               & 0.0819          & 0.3832          & 0.4650          \\
for graduate                       & 0.0858          & 0.4009          & 0.4866          \\
Pivot SDA AI   \& Autonomy Sandbox & 0.1299          & 0.6361          & 0.7659          \\
ItTakesTwoToTango                  & 0.0800          & 0.6922          & 0.7721          \\
chusunhao                          & 0.0584          & 0.0584          & 0.8151         

\end{tabular}%
}
\caption{Summary of the SPEC2021 results for the Sunlamp Dataset.}
\label{tab:leaderboard_sunlamp}
\end{table}

% \usepackage{graphicx}
\begin{table}[htbp]
\centering
\resizebox{0.65\textwidth}{!}{%
\begin{tabular}{l|cc|c}
\multicolumn{1}{c|}{\textbf{Team Name}} & \textbf{$S^j_{pos}$} & \textbf{$S^j_{ori}$} & \textbf{ $S^j_{total}$} \\ \hline
VPU (Ours)                       & 0.0166          & \textbf{0.0646} & \textbf{0.0812 } \\
TangoUnchained                   & \textbf{0.0164} & 0.0653          & 0.0816          \\
lava1302                         & 0.0298          & 0.0820          & 0.1117          \\
haoranhuang\_njust               & 0.0456          & 0.1443          & 0.1742         \\
u3s\_lab                         & 0.0434          & 0.1391          & 0.1824         \\
for   graduate                   & 0.0805          & 0.4070          & 0.4874          \\
bbnc                             & 0.0879          & 0.4088          & 0.4981          \\
chusunhao	                     & 0.0456	       & 0.17215	     & 0.56685       \\
Pivot SDA AI \& Autonomy Sandbox & 0.0999          & 0.5268          & 0.6266         \\
ItTakesTwoToTango 	             & 0.0811          & 0.61745         & 0.69845
\end{tabular}%
}
\caption{Summary of the SPEC2021 results averaged over both Datasets.}
\label{tab:leaderboard_average}
\end{table}

% FIUGRES -----------------------
\begin{figure}[htbp]
\begin{center}
   \includegraphics[width=0.7\linewidth]{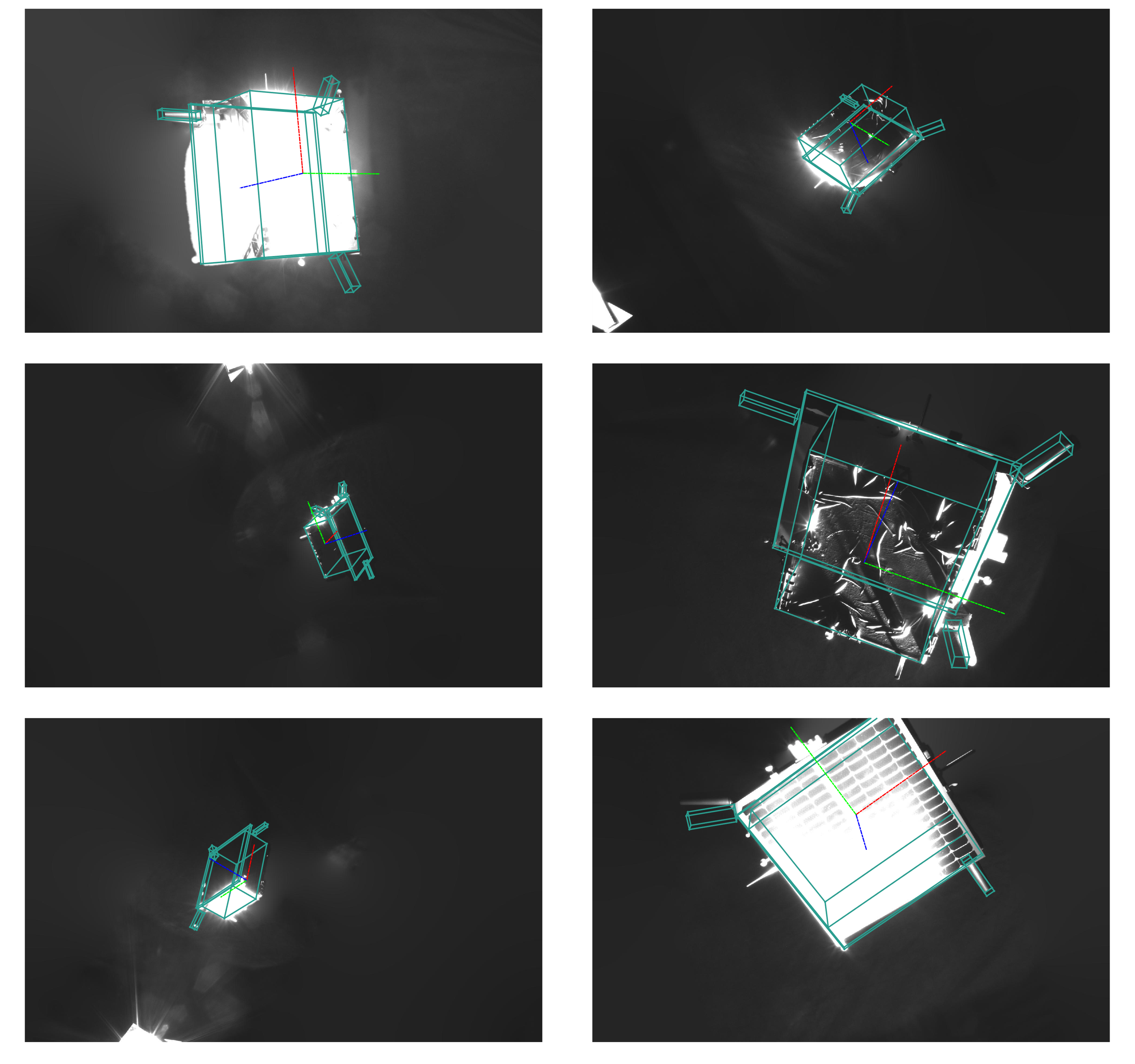}
\end{center}
   \caption{Example results of the proposed algorithm on the Sunlamp test dataset. The estimated pose is represented with a wire-frame model over the image.}
\label{fig:sunlamp_examples}
\end{figure}

\begin{figure}[h]
\begin{center}
   \includegraphics[width=0.7\linewidth]{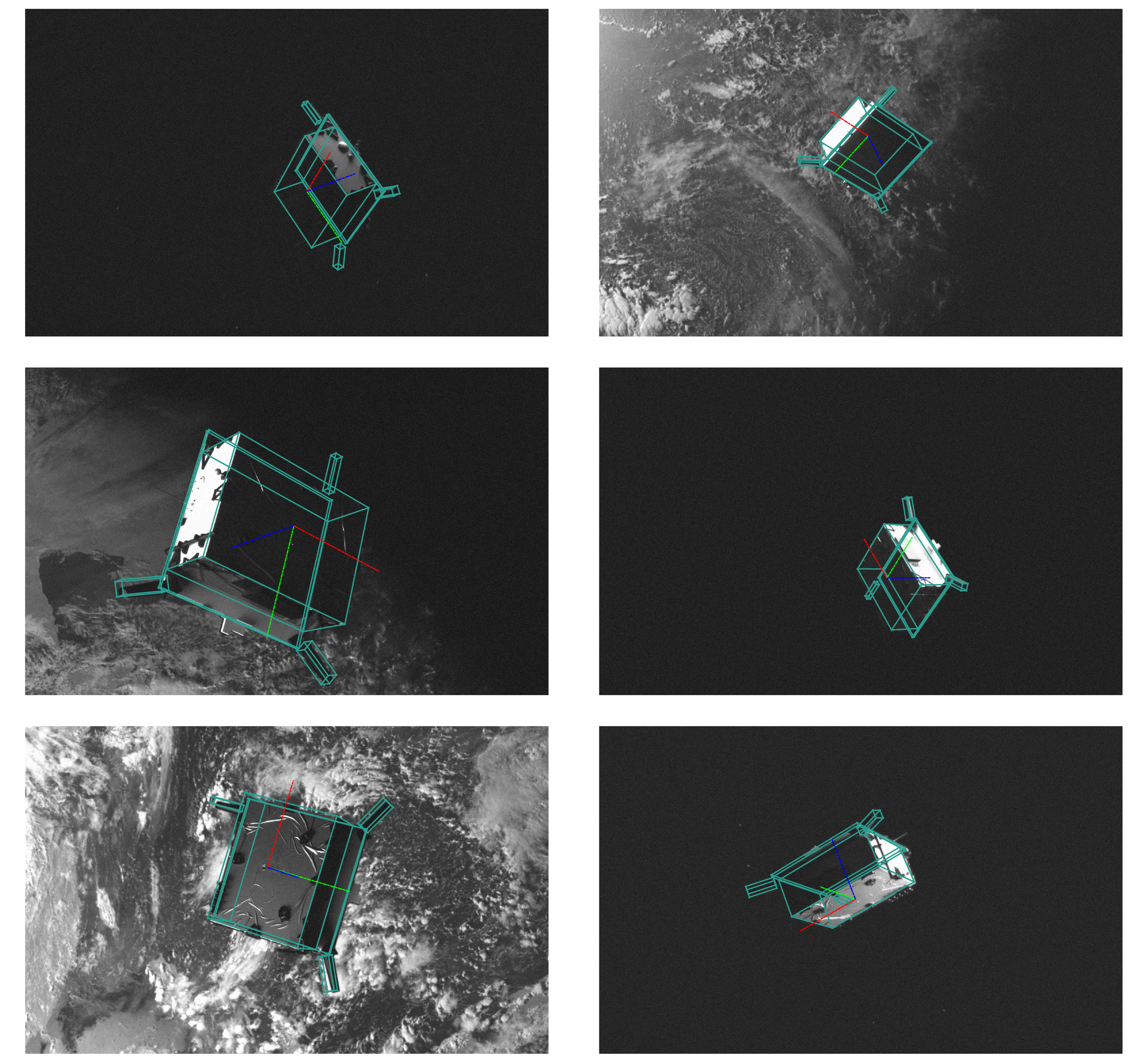}
\end{center}
   \caption{Example results of the proposed algorithm on the Lightbox test dataset. The estimated pose is represented with a wire-frame model over the image.}
\label{fig:lightbox_examples}
\end{figure}

For the Challenge, we trained the model described in Section \ref{section:PoseEstimation} during 35 epochs on the Synthetic dataset at a resolution of 640x640 pixels. Then, we trained individual models following Algorithm \ref{alg:algorithm1} for the Sunlamp and Lightbox target datasets during 50 iterations. Fig. \ref{fig:sunlamp_examples} and Fig. \ref{fig:lightbox_examples} provide visual results of the estimated poses for the Sunlamp and Lightbox datasets. The estimated poses are represented with the 3-axis plot and the wireframe model projected over the image. They show that the proposed method can handle strong reflections, presence of other elements on the image, complex background and close views of the object.

\subsection{Additional Experiments} \label{seq:additional_experiments}

\begin{figure}[]
\centering
  \includegraphics[width=\linewidth]{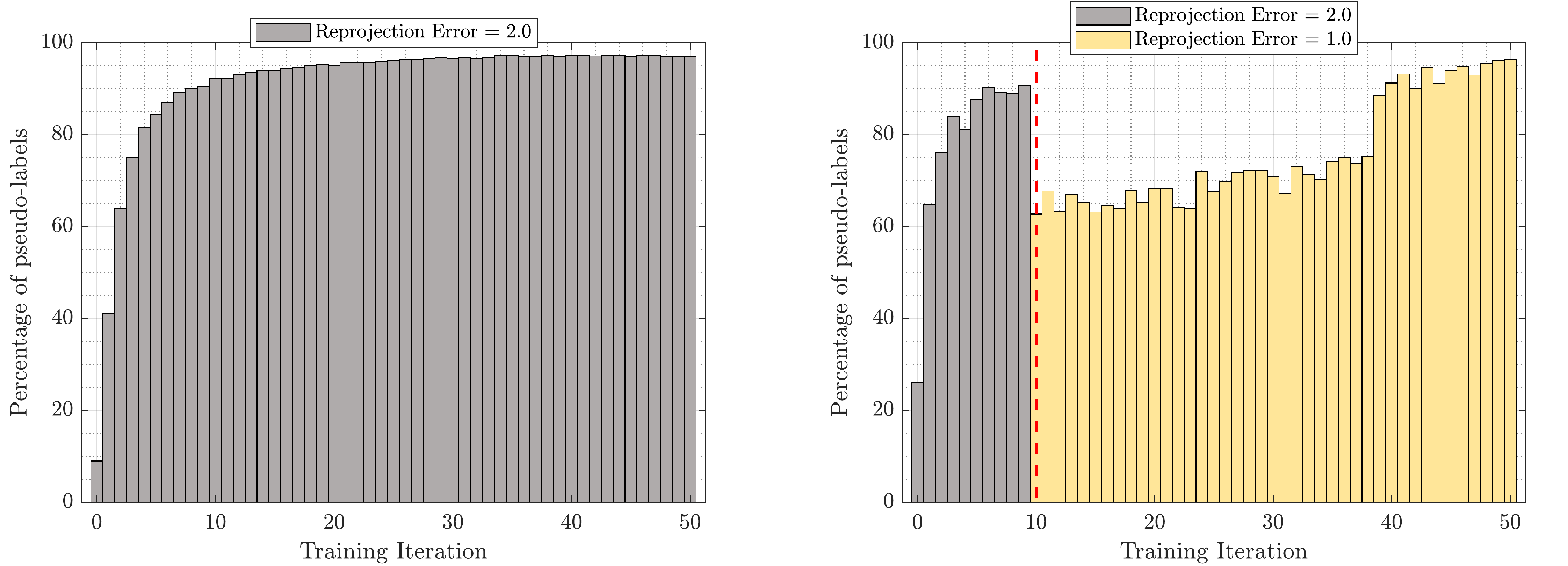}
  \caption{Percentage of pseudo-labels created in the dataset as a function of the training iterations over Algorithm \ref{alg:algorithm1}. Left plot represents the pseudo-labels over the Sunlamp dataset for a constant reprojection error constraint of 2.0 for the RANSAC EPnP algorithm. Right plot represents the pseudo-labels over the Lightbox dataset for a variable reprojection error that starts at 2.0 and drops to 1.0 after 10 iterations.}\label{fig:pseudo-labels}
\end{figure}

We show in Fig. \ref{fig:pseudo-labels} the effect of iterating over Algorithm \ref{alg:algorithm1} on the number of generated pseudo-labels. The left plot corresponds to the strategy followed over the Sunlamp dataset, where the reprojection error that controls the convergence criteria of RANSAC-EPnP remains fixed during all iterations with a value of 2.0. It can be observed that the percentage of pseudo-labels reaches 90\% after 10 iterations and slowly increases from there, suggesting a possible overfit to wrongly estimated labels. The right plot corresponds to the Ligthbox dataset, where a different approach was followed. Motivated by the fact that the Lightbox test set is more visually similar to the training set than the Sunlamp test set, we hypothesize that restricting the label generation by reducing the reprojection error would lead to better pseudo-labels and hence better perfromance. To test this, the reprojection error was decreased after 10 iterations (red dashed line) to 1.0. It can be observed that the percentage of pseudo-labels quickly drops and then grows at a slower rate. However, based on the final results we cannot conclude that this strategy improved the results compared to using a fixed reprojection error, as the achieved errors for both test sets are in a similar range. 

To show the effectiveness of the proposed method on the target test sets, we evaluated the proposed pose estimation algorithm against a baseline considering only the 2D heatmap loss. The models were trained on the source synthetic dataset during 10 epochs and tested over a pseudo-test dataset. The pseudo-test set is obtained by running Algorithm \ref{alg:algorithm1} until approximately 30 \% of the target dataset is labelled. This number is chosen as a compromise between the necessary number of pseudo-labels to obtain a representative analysis and the quality of the labels obtained. The pseudo-test set approach is followed as the real labels of the test set are not made publicy available in the challenge. The results are shown in Table \ref{tab:contribution} with the validation set from the source dataset as reference. It can be observed that the proposed method, despite obtaining similar values on the validation (Source set) is able to improve the results on the target (test) sets.

% \usepackage{multirow}
% \usepackage{graphicx}
\begin{table}[h!]
\resizebox{\textwidth}{!}{%
\begin{tabular}{l|ccc|ccc|ccc} 
\multicolumn{1}{c|}{\multirow{2}{*}{Method}} & \multicolumn{3}{c|}{\textbf{Sunlamp}}                                                         & \multicolumn{3}{c|}{\textbf{Lightbox}}                                                        & \multicolumn{3}{c}{\textbf{Validation}}                                                                    \\ \cline{2-10} 
\multicolumn{1}{c|}{}                        & \multicolumn{1}{c|}{$S^j_{pos}$}     & \multicolumn{1}{c|}{$S^j_{ori}$}     & $S^j_{total}$   & \multicolumn{1}{c|}{$S^j_{pos}$}     & \multicolumn{1}{c|}{$S^j_{ori}$}     & $S^j_{total}$   & \multicolumn{1}{c|}{$S^j_{pos}$}    & \multicolumn{1}{c|}{$S^j_{ori}$}     & $S^j_{total}$                 \\ \hline
Baseline                                     & \multicolumn{1}{c|}{0.3095}          & \multicolumn{1}{c|}{0.9229}          & 1.2324          & \multicolumn{1}{c|}{0.2888}          & \multicolumn{1}{c|}{0.8935}          & 1.1823          & \multicolumn{1}{c|}{\textbf{0.0086}}              & \multicolumn{1}{c|}{0.02627}     & \textbf{0.03490}               \\
Proposed                                     & \multicolumn{1}{c|}{\textbf{0.2138}} & \multicolumn{1}{c|}{\textbf{0.8554}} & \textbf{1.0692} & \multicolumn{1}{c|}{\textbf{0.2837}} & \multicolumn{1}{c|}{\textbf{0.7709}} & \textbf{1.0546} & \multicolumn{1}{c|}{0.0088} & \multicolumn{1}{c|}{\textbf{0.0256}} & 0.0355
\end{tabular}%
}\caption{Influence of the proposed pose estimation method on the position, orientation and total score on a pseudo-test set and on the validation set.}\label{tab:contribution}
\end{table}

\section{Conclusions}
We have presented a method for spacecraft pose estimation that integrates structure information in the learning process by incorporating pose and 3D alignment information in a combined loss function. We argue that incorporating such information provides robustness to domain shifts when the imaged objects keep the same structure although they are captured under different illumination conditions. In addition we have introduced a method for pseudo-labelling in pose estimation that exploits the consensus within a network and provides robustness to label noise. Our solution has ranked second on both datasets for the SPEC2021 Challenge, achieving the best average score.
\subsubsection{Acknowledgements.} This work is supported by Comunidad Autónoma de Madrid (Spain) under the Grant IND2020/TIC-17515.

\clearpage


\bibliographystyle{splncs04}
\bibliography{177}
\end{document}